\documentclass[conference]{IEEEtran}
\IEEEoverridecommandlockouts
\usepackage{cite}
\usepackage{amsmath,amssymb,amsfonts}
\usepackage{algorithmic}
\usepackage{comment}
\usepackage{graphicx}
\usepackage{algorithm}
\usepackage{textcomp}
\usepackage{xcolor}
\usepackage{graphicx}
\usepackage{xcolor}

\newtheorem{assumption}{Assumption}
\newtheorem{proposition}{Proposition}
\def\BibTeX{{\rm B\kern-.05em{\sc i\kern-.025em b}\kern-.08em
    T\kern-.1667em\lower.7ex\hbox{E}\kern-.125emX}}
\begin{document}

\title{An Exact Branch-Price-and-Cut Algorithm for the Vehicle Routing Problem with Floating Targets}

\author{\IEEEauthorblockN{Micha\l{} Paradowski\IEEEauthorrefmark{1},
Anoop Bhat\IEEEauthorrefmark{2},
Zhongqiang Ren\IEEEauthorrefmark{3} and
Howie Choset\IEEEauthorrefmark{2}}
\IEEEauthorblockA{\IEEEauthorrefmark{1}\textit{Institute of Control and Computation Engineering, Warsaw University of Technology}, Warsaw, Poland}
\IEEEauthorblockA{\IEEEauthorrefmark{2}\textit{Robotics Institute, Carnegie Mellon University}, Pittsburgh, PA, USA}
\IEEEauthorblockA{\IEEEauthorrefmark{3}\textit{Global College, Shanghai Jiao Tong University}, Shanghai, China}
\IEEEauthorblockA{michal.paradowski.stud@pw.edu.pl, \{agbhat, choset\}@andrew.cmu.edu, zhongqiang.ren@sjtu.edu.cn}}

\maketitle

\begin{abstract}
The Vehicle Routing Problem with Floating Targets (VRPFT) is a generalization
of the capacitated vehicle routing problem where targets may actively move to
locations ``convenient'' for the vehicles. In particular, each target may
leave its home location at a bounded speed. A solution to the VRPFT is a
sequence of targets for each vehicle to visit and a meeting point in
space-time for each target. In this paper, we solve the VRPFT exactly with a
branch-price-and-cut algorithm. Our key contribution is a dominance criterion
for the pricing labeling algorithm that utilizes closed-form lower and upper
bounds on a vehicle's trajectory cost given a fixed sequence of targets. We prove that the criterion preserves
optimality, and the algorithm has smaller computation time than the exact dynamic program of prior work
on identical hardware. We also provide a relaxed criterion that prunes more
aggressively; the resulting heuristic is faster than our exact algorithm and matched every certified
optimum on the benchmark.
\end{abstract}

\section{Introduction}
\label{sec:intro}
The vehicle routing problem (VRP), introduced by Dantzig and Ramser \cite{b9}, asks
for a set of minimum-cost vehicle paths that serve a set of static
targets subject to capacity constraints. It is among the most
studied problems in operations research, both for its practical importance in
transportation and logistics and for its computational difficulty \cite{b10}. A
defining assumption of the classical VRP and most of its variants is that each
target occupies a fixed, known location that a vehicle must visit.
\begin{figure}[t]
\centering
\includegraphics[width=\columnwidth]{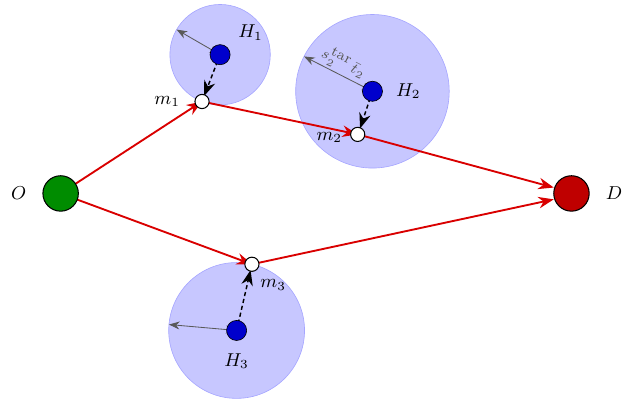}
\caption{A VRPFT instance with two vehicles. Each target $j$ leaves its home
$H_{j}$ at time $t=0$ at a speed of at most $s^{\textnormal{tar}}_{j}$, so it can be anywhere in the disk of radius $s^{\textnormal{tar}}_{j}\bar t_{j}$ at the time
$\bar t_{j}$ of its meeting;
the disk grows with the elapsed time, so targets served later can
travel farther. Targets~1 and~3 walk to the boundary of their disks, whereas
target~2 does not need its full range.}
\label{fig:instance}
\end{figure}
This assumption does not hold in a growing class of applications in which the targets are themselves mobile and can cooperate with the fleet. In
ride-sharing and last-mile delivery, a passenger may walk a short distance to meet
a vehicle at a more convenient point than their home \cite{b23,b24,b13}; in aerial refueling, a target and an interceptor can jointly choose a rendezvous \cite{b25}. 

Motivated
by such settings, Gambella \textit{et al.} \cite{b8} introduced the vehicle routing
problem with floating targets (VRPFT), a generalization of the VRP in which each
target may leave its home location and travel, up to a maximum speed, toward a selected meeting
point. Because a target can move in any direction, the set of locations at which it can be picked up by a given time is a disk
rather than a single point (Fig.~\ref{fig:instance}).

Solving the VRPFT to optimality is challenging due to the coupling between the discrete routing decisions and continuous meeting-point decisions. In this paper, we solve the VRPFT exactly with a branch-price-and-cut algorithm. Central to the algorithm is a dominance criterion for the pricing labeling algorithm, built on closed-form lower and upper bounds on the minimum time required for a vehicle to meet a fixed sequence of targets. The criterion is provably optimality-preserving, and the resulting algorithm runs faster than the exact dynamic program of \cite{b13} on identical hardware. Relaxing the criterion yields a heuristic that prunes more aggressively and is faster than our exact algorithm by a factor ranging from 1.3 to 6.2 on the benchmark while matching every certified optimum.

The remainder of this paper is organized as follows. Section~II reviews prior work and Section~III states the VRPFT; the remaining
sections describe the branch-price-and-cut algorithm and report the
computational results.

\section{Prior Work}
\label{sec:previous}

The VRPFT was introduced by Gambella \textit{et al.} \cite{b8}, who provided a branch-and-price algorithm for the problem as well as benchmark instances that we use in this paper. Several
of the instances in \cite{b8} with relatively more targets and larger capacity remained unsolved within their two-hour time limit.

Zhang \textit{et al.} \cite{b13} provide an exact dynamic programming
algorithm for the VRPFT that matches the published optima on the benchmark of
\cite{b8} at a fraction of the running time of the algorithm from \cite{b8}; a heuristic acceleration also
improves on the heuristic of \cite{b8} on the instances where \cite{b8} did not find the optimum.

Wang \textit{et al.}~\cite{b41} studied a drone shore-to-ship
delivery problem in which vessels sail toward drone meeting points,
generalizing the VRPFT with payload capacities, time windows, and a non-linear
energy budget. They solve it by branch-and-price with a novel primal heuristic, and also report results on the benchmark of \cite{b8}. Their pricing
problem couples sequencing with continuous meeting points as ours does, but is
handed to a commercial Mixed-Integer Second-Order Cone Programming (MISOCP) solver.

The VRPFT belongs to a broader family of routing problems with mobile or
spatially flexible meeting points. For example, the moving-target traveling
salesman problem \cite{b15} and its multi-vehicle extensions
\cite{b39,b40} require intercepting targets that follow known trajectories, whereas in the VRPFT, the trajectories of the targets and vehicles are planned jointly. Additionally, in the traveling salesman problem with neighborhoods
\cite{b43,b44}, each customer may be served anywhere within a fixed region,
which is a disk in the close-enough variant \cite{b16}. In the VRPFT, that disk
instead grows over time as the target walks from its home location.

Closest to our approach, Bhat \textit{et al.} \cite{b14} proposed
an exact branch-and-price algorithm for the moving-target VRP, in which targets moving along fixed trajectories
must be intercepted within bounded time windows. Their contribution is a labeling
algorithm with a dominance criterion based on relaxing the continuity constraints of an agent's trajectory, building on
the relaxed-continuity ideas of \cite{b39}. This dominance test
is the inspiration for the dominance criterion used in our work, but we handle the scenario where the targets' trajectories are not fixed, unlike \cite{b14}. A structural difference is worth emphasizing: the bounded time windows of the
moving-target VRP \cite{b14} make many target-to-target transitions infeasible, allowing the pruning of many target sequences. Meanwhile, in the VRPFT, every unvisited target
remains reachable at any time, so no such pruning is available.

Finally, our algorithm builds on advances in routing for static targets. In particular, we adapt rounded-capacity cuts \cite{b19} and subset-row
cuts \cite{b20} to the floating-target setting.
\section{Problem Statement and Formulation}
\label{sec:problem}
\subsection{Problem Description}
A fleet of $K$ identical vehicles, each with maximum speed $s^{\textnormal{veh}}$,
must serve a set of $n$ targets. A single vehicle can serve at most $Q$ targets, where $Q$ is called the \emph{capacity}. All used vehicles depart from a common origin
$O \in \mathbb{R}^{2}$ at time $t=0$. Not all vehicles must be used, but if a vehicle is used, it must terminate at a destination
$D \in \mathbb{R}^{2}$, common to all vehicles. Each target $j$ has a home location $H_{j} \in \mathbb{R}^{2}$
and a maximum speed $s^{\textnormal{tar}}_{j}$. At time $t=0$ every target is located at its home;
thereafter it may travel in any direction at a speed no greater than $s^{\textnormal{tar}}_{j}$.
Throughout the paper we make the following standing assumption.
\begin{assumption}
\label{ass:speed}
No target's speed limit is larger than the vehicles' speed limit: $s^{\textnormal{veh}}\ge s^{\textnormal{tar}}_{j}$ for all
$j=1,\dots,n$.
\end{assumption}
Assumption~\ref{ass:speed} holds for all test instances proposed by Gambella \textit{et al.}
and is sensible in the context of applications listed in Section~\ref{sec:intro}.

Following
\cite{b8}, the origin is treated as the fictitious target $j=0$ and the destination as
the fictitious target $j=n+1$. A \textit{route} is a sequence
\begin{equation}
r=(0,\,j_{1},\,j_{2},\,\dots,\,j_{p},\,n{+}1)
\end{equation}
that starts at the origin $O$, visits $1 \le p \le Q$ pairwise distinct targets, and ends at
the destination $D$.

The VRPFT seeks vehicle routes visiting each target exactly once, together with
a meeting point in space and a meeting time for each target, subject to the vehicle capacities and speed limits of vehicles and targets. The objective is to minimize the sum of vehicles' travel times.

\subsection{Notation}
$\lVert\cdot\rVert$ refers to Euclidean distance, i.e. $\lVert\cdot\rVert_{2}$. Let $\Omega$ denote the set of all routes. Because the
vehicles are identical, $\Omega$ does not depend on the vehicle index. A \emph{partial route} is a prefix of a route that stops short of the
destination,
\begin{equation}
\Gamma = (0, j_1, j_2, \dots, j_q), \qquad 0 \le q \le Q,
\end{equation}
visiting $q$ pairwise distinct targets.
An \textit{arc} $e_{ij}$ is an ordered pair of targets $i$ and $j$. We say that a route $r$ \textit{uses} an arc $e_{ij}$ if, and only if, $(i, j)$ is a contiguous subsequence of $r$. We denote this by $e_{ij}\sqsubseteq r$. We define the set of all arcs as
\begin{equation}
\mathcal{E}=\{e_{ij}: i, j\in\{0, 1, ..., n+1\}, i\neq j\}
\end{equation}

\subsection{Route Costs}
\label{sec:reformulation}

For a route $r$, the visiting order is fixed, so the remaining decisions are purely
continuous: where each meeting takes place and when. Let $m_{i}$ be the $i$-th
meeting point. Let $t_i$ be the time elapsed between the $(i{-}1)$-th and the $i$-th meeting,
$i=1,\dots,p+1$. The cost of
$r$ is the optimal value of
\begin{align}
c_{r}=\min\ & \sum_{i=1}^{p+1} t_{i} \label{eq:routecost}\\
\text{s.t. } & \lVert m_{i}-m_{i-1}\rVert \le s^{\textnormal{veh}} t_{i}, && i=1,\dots,p{+}1, \label{eq:rc_veh}\\
& \lVert m_{i}-H_{j_{i}}\rVert \le s^{\textnormal{tar}}_{j_{i}} \sum_{i'=1}^{i} t_{i'}, && i=1,\dots,p, \label{eq:rc_tgt}\\
& m_{0}=O, \quad m_{p+1}=D, \quad t_{i}\ge 0. \label{eq:rc_bnd}
\end{align}
The objective \eqref{eq:routecost} is to minimize the total elapsed time. \eqref{eq:rc_veh} enforces the vehicle's speed limit. \eqref{eq:rc_tgt} enforces the targets' speed limit. \eqref{eq:rc_bnd} ensures that the vehicle starts by ``meeting" the fictitious target at the vehicle's home location, and ends by ``meeting" the fictitious target at the vehicle's destination. Constraints \eqref{eq:rc_veh} and \eqref{eq:rc_tgt} are second-order cones and the
objective is linear, so \eqref{eq:routecost}--\eqref{eq:rc_bnd} is a \textit{convex}
second-order cone program, solvable to global optimality.

A property of \eqref{eq:routecost}--\eqref{eq:rc_bnd} is that
deleting a target from a route never increases its cost, by the following argument. Let $r'$ be the
route obtained by removing the $k$-th target from $r$, and let $\{(m_i, t_i)\}_{i = 1}^{p}$
be an optimal solution of \eqref{eq:routecost}--\eqref{eq:rc_bnd} for $r$.
Keep the meeting point and elapsed time of every remaining target, and let
the vehicle travel from $m_{k-1}$ straight to $m_{k+1}$ over the combined
time $t_k+t_{k+1}$: by the triangle inequality and \eqref{eq:rc_veh},
$\lVert m_{k+1}-m_{k-1}\rVert\le\lVert m_{k}-m_{k-1}\rVert
+\lVert m_{k+1}-m_{k}\rVert\le s^{\textnormal{veh}}(t_k+t_{k+1})$, so the leg is feasible,
with the vehicle waiting out any slack. The elapsed time at every surviving
meeting is unchanged, so \eqref{eq:rc_tgt} still holds, and the total time
equals $c_r$; hence $c_{r'}\le c_{r}$.

\subsection{Integer Linear Problem for VRPFT}
Let $a_{jr}\in\{0,1\}$ indicate whether route $r$ serves target $j$, and let the
route-usage variable $\theta_{r}$ be $1$ if route $r$ is operated by some vehicle and
$0$ otherwise. The VRPFT is then equivalent to the integer linear problem (ILP)
\begin{align}
\quad \min\ & \sum_{r\in\Omega} c_{r}\,\theta_{r} \label{eq:mp_obj}\\
\text{s.t. } & \sum_{r\in\Omega} a_{jr}\,\theta_{r} \ge 1, && j=1,\dots,n, \label{eq:mp_cover}\\
& \sum_{r\in\Omega} \theta_{r} \le K, \label{eq:mp_fleet}\\
& \theta_{r}\in\{0,1\}, && r\in\Omega. \label{eq:mp_bin}
\end{align}
The objective \eqref{eq:mp_obj} minimizes the sum of used route costs. Constraints \eqref{eq:mp_cover} require every target to be served, and
\eqref{eq:mp_fleet} limits the number of operated routes to the fleet size. \eqref{eq:mp_bin} requires all variables to be binary.

\subsection{LP relaxation of the ILP}
\label{sec:relaxation}
We define the Master Problem [MP] as a LP relaxation of the ILP analogously to \cite{b14}. The convex
relaxation of ILP \eqref{eq:mp_obj}--\eqref{eq:mp_bin} consists of
\eqref{eq:mp_obj}--\eqref{eq:mp_fleet} with \eqref{eq:mp_bin} replaced by
$\theta_r \ge 0$. An optimal solution $\theta^{*}$ of the relaxation may be
fractional, i.e. non-integer. The branch-and-bound therefore iteratively breaks ILP
\eqref{eq:mp_obj}--\eqref{eq:mp_bin} into subproblems, each disallowing a set
of arcs $\mathcal{F} \subset \mathcal{E}$ to eliminate fractional solutions.
We define ILP-$\mathcal{F}$ as ILP \eqref{eq:mp_obj}--\eqref{eq:mp_bin} with
the additional constraints
\begin{align}
\theta_{r} = 0, \qquad r \in \bigl\{\, r' \in \Omega \;:\; \exists\, e \in \mathcal{F},\ e \sqsubseteq r' \,\bigr\}.
\label{eq:rmp_F}
\end{align}
and LP-$\mathcal{F}$ as the LP relaxation of ILP-$\mathcal{F}$, obtained
as above.

The set $\Omega$ grows factorially with $n$, so LP-$\mathcal{F}$
cannot be written out explicitly. It is therefore solved by lazy route
generation (often referred to as column generation in the literature) \cite{b12}. This procedure begins with a small subset of all routes, $\Omega'$, and enlarges $\Omega'$ as needed. Thus, the restricted master problem, RMP-$\mathcal{F}$, is obtained from \eqref{eq:mp_obj}--\eqref{eq:rmp_F}, by replacing \eqref{eq:mp_bin} with
\begin{equation}
    \theta_r \geq 0, r\in\Omega'.
\end{equation}
and taking every sum over a pool $\Omega'\subseteq\Omega$. 
\section{The Branch-Price-and-Cut Algorithm}
\label{sec:bprc}

\subsection{Overview}
\label{sec:bprc_overview}
We solve \eqref{eq:mp_obj}--\eqref{eq:mp_bin} by branch-price-and-cut
\cite{b11}, summarized in Algorithm~\ref{alg:bpc}. Alg. \ref{alg:bpc} searches a tree where each node is a subproblem ILP-$\mathcal{F}$ of Section~\ref{sec:relaxation}:
node $N$ is identified with its forbidden arcs $\mathcal{F}_N$, the root is
$\mathcal{F}_0=\emptyset$ (line~\ref{ln:root}), and the two children of $N$
are obtained by adding arcs to $\mathcal{F}_N$ in a way that eliminates fractional solutions from LP-$\mathcal{F}_N$
(lines~\ref{ln:branch_start}--\ref{ln:branch_end}).
An exhaustive backtracking search (Appendix~\ref{app:backtracking}), in the spirit of the depth-first
feasible-solution construction of \cite{b14}, builds an initial feasible
solution (one exists iff $KQ\ge n$), supplying the first routes and the
incumbent, i.e.\ the best feasible solution known so far
(lines~\ref{ln:init_start}--\ref{ln:init_end}).
This backtracking search is reused when a node's
pool cannot cover every target under $\mathcal{F}_N$
(lines~\ref{ln:repair_start}--\ref{ln:repair_end}).

At every node,
LP-$\mathcal{F}_N$ is solved by lazy route generation over the pool
$\Omega'$. That is, we first solve RMP-$\mathcal{F}_N$ (lines~\ref{ln:cg_start}--\ref{ln:cg_end};
Section~\ref{sec:bprc_cg}), then search for routes to improve the RMP solution by solving a pricing problem
(Sections~\ref{sec:bprc_labeling}--\ref{sec:bprc_dominance}). We alternate between the RMP and pricing problem until we find no new
routes in the pricing problem
(Section~\ref{sec:bprc_socp}). At the root, route generation alternates with the \emph{separation} of valid inequalities, i.e.,
the search for a member of a given family that the current relaxation solution violates
(lines~\ref{ln:cut_start}--\ref{ln:cut_end}; Section~\ref{sec:bprc_cuts}),
constraints that cut off fractional solutions while keeping an optimal integer one. These valid inequalities remain
in force throughout the tree. Once route generation converges,
the optimal cost of the RMP is a lower bound on the optimal integer solution for LP-$\mathcal{F}$ (line~\ref{ln:bound}). Then, we solve a version of the RMP where $\theta_r$ values are constrained to be integer, and if we obtain a better solution than the incumbent, we update the incumbent
(lines~\ref{ln:ip_start}--\ref{ln:ip_end}). Next, the node is pruned if its bound
is no better than the incumbent or its relaxation is integral
(lines~\ref{ln:prune_start}--\ref{ln:prune_end}); otherwise we branch
(lines~\ref{ln:branch_start}--\ref{ln:branch_end}).

We branch on arc flows. The flow of arc $(u,v)$ is
$x_{uv}=\sum_{r} b_{uv,r}\,\theta_{r}$, where $b_{uv,r}=1$ iff $v$ immediately
follows $u$ in route $r$. Branching on an arc $(u,v)$ with fractional flow
creates two children: child $0$ forbids $(u,v)$, and child $1$ forces it by
forbidding every other arc leaving $u$ (unless $u=0$) or entering $v$ (unless
$v=n{+}1$). Routes violating a node's forbidden arcs are disabled in the
restricted master, and the labeling algorithm of
Section~\ref{sec:bprc_labeling} never extends along forbidden arcs. The two
children partition the integer solutions of the parent, so no integer solution
is lost.

The branching arc is chosen by strong branching \cite{b11}. Let $z^{k}_{uv}$,
$k\in\{0,1\}$, be the lower bound of child $k$ if we branch on $(u,v)$. For
each fractional arc we compute an estimate $\tilde z^{\,k}_{uv}$ of
$z^{k}_{uv}$ by re-solving the restricted master (with the cuts retained) over
$\bar\Omega^{k}_{uv}$, the routes of the current $\Omega'$ that child $k$
allows, \emph{without} invoking pricing; $\tilde z^{\,k}_{uv}=+\infty$ if
$\bar\Omega^{k}_{uv}$ does not cover every target. We branch on the arc that
maximizes the weaker child's estimate,
\begin{equation}
(u^{*},v^{*})\;\in\;
\operatorname*{arg\,max}_{(u,v):\,0<x_{uv}<1}\;
\min\bigl\{\tilde z^{\,0}_{uv},\,\tilde z^{\,1}_{uv}\bigr\},
\label{eq:sb_maxmin}
\end{equation}
with ties broken toward the most fractional flow,
$\min\,\lvert x_{uv}-\tfrac12\rvert$. Since pricing is skipped,
$\tilde z^{\,k}_{uv}$ only \emph{over}estimates $z^{k}_{uv}$; the estimates
rank candidate arcs and are never used for pruning, so they affect the quality
of the selection, not the correctness of the branching scheme.

On termination, repeated targets are stripped from the routes of $x^{*}$, which does not increase its cost (Section~\ref{sec:reformulation}).
Algorithm~\ref{alg:bpc} is exact (Proposition~\ref{prop:exact}). Section~\ref{sec:bprc_relaxed} proposes a relaxed dominance criterion that trades exactness for speed; Section~\ref{sec:results} evaluates both.

\begin{algorithm}[t]
\caption{Branch-price-and-cut for the VRPFT}
\label{alg:bpc}
\begin{algorithmic}[1]
\STATE $\Omega' \gets$ feasible solution found by exhaustive backtracking \label{ln:init_start}
\STATE $(z^{*}, x^{*}) \gets$ cost and routes of that solution \label{ln:init_end}
\STATE stack $\gets$ \{root node $0$: $\mathcal{F}_0=\emptyset$\} \label{ln:root}
\WHILE{stack $\neq \emptyset$}
  \STATE pop node $N$; disable routes using $\mathcal{F}_N$ \label{ln:pop}
  \IF{restricted master is infeasible under $\mathcal{F}_N$} \label{ln:repair_start}
    \STATE rebuild a set of routes respecting $\mathcal{F}_N$ by backtracking
    \IF{none exists}
      \STATE prune $N$
    \ELSE
      \STATE add its routes, updating $(z^{*},x^{*})$ if cheaper \label{ln:repair_end}
    \ENDIF
  \ENDIF
  \REPEAT
    \REPEAT \label{ln:cg_start}
      \STATE solve RMP-$\mathcal{F}_N$ with the cuts in $\mathcal{S},\mathcal{C}$; collect duals
      \STATE add every route with $\bar{c}_{r}<-\varepsilon$ \eqref{eq:redcost} found by the labeling algorithm
    \UNTIL{no route was added} \label{ln:cg_end}
    \IF[separate valid inequalities]{$N$ is the root} \label{ln:cut_start}
      \STATE separate RCCs \eqref{eq:rcc}
      \IF{no RCC was violated \AND the relaxation is fractional}
        \STATE separate SRCs \eqref{eq:src} \label{ln:cut_end}
      \ENDIF
    \ENDIF
  \UNTIL{no cut was added}
  \STATE $z_{N} \gets$ converged relaxation value \COMMENT{bound at $N$} \label{ln:bound}
  \STATE solve the master as a $0/1$ program over $\Omega'$ \label{ln:ip_start}
  \STATE update $(z^{*},x^{*})$ \label{ln:ip_end}
  \IF{$z_{N} \ge z^{*}$} \label{ln:prune_start}
    \STATE prune $N$
  \ELSIF{the relaxation solution is integral}
    \STATE $(z^{*},x^{*}) \gets$ relaxation solution; prune $N$ \label{ln:prune_end}
  \ELSE
    \STATE choose a fractional arc $(u,v)$ by strong branching \label{ln:branch_start}
    \STATE push the child forbidding $(u,v)$ and the child forcing $(u,v)$ \label{ln:branch_end}
  \ENDIF
\ENDWHILE
\RETURN $x^{*}$, with objective $z^{*}$
\end{algorithmic}
\end{algorithm}

\subsection{Valid Inequalities}
\label{sec:bprc_cuts}
The solution to the master problem is in general fractional. We tighten the relaxation with two
standard families of cuts, rounded capacity cuts (RCCs) \cite{b19} and subset-row
cuts (SRCs) \cite{b20}. Throughout, $\mathcal{S}$ and
$\mathcal{C}$ denote the sets of rounded capacity and subset-row cuts
currently in the master.

\subsubsection{Rounded capacity cuts}
For a target set $S\subseteq\{1,\dots,n\}$ with $|S|>Q$, let
$\partial S=\{e_{ij}\in \mathcal{E}:\ \lvert\{i,j\}\cap S\rvert=1\}$ be the set of arcs
crossing the boundary of $S$, and let
$\chi_{r}(S)=\sum_{e\sqsubseteq r}\mathbf{1}[e\in\partial S]$ count the
crossings for route $r$. The rounded capacity cut (RCC) in its cutset form
\cite{b19} is
\begin{equation}
\sum_{r\in\Omega}\chi_{r}(S)\,\theta_{r}\ \ge\ 2\,\lceil |S|/Q\rceil.
\label{eq:rcc}
\end{equation}
Its validity in the capacitated VRP \cite{b19} relies only on every route
beginning and ending outside $S$ and serving at most $Q$
targets---both true for the VRPFT.

\subsubsection{Subset-row cuts}
For a triple $C\subseteq\{1,\dots,n\}$, $|C|=3$, a solution visiting each target once (without loss of optimality by Section~\ref{sec:reformulation}) can contain at most one route that serves two or more targets of $C$. This gives the subset-row cut (SRC)~\cite{b20}
\begin{equation}
\sum_{r\in\Omega}\Big\lfloor\tfrac{1}{2}\textstyle\sum_{j\in C} a_{jr}\Big\rfloor\,
\theta_{r}\ \le\ 1.
\label{eq:src}
\end{equation}
Both the RCC and SRC families are separated exactly, by enumeration over all sets $S$ with
$|S|>Q$ and all $\binom{n}{3}$ triples, most violated first, only at the root
node and only once pricing has converged. Each separated cut is appended to
LP-$\mathcal{F}$ and to RMP-$\mathcal{F}$, with its sum restricted to $\Omega'$ in the
latter; the cuts then remain in the master throughout the tree, so
$\mathcal{S}$ and $\mathcal{C}$ are fixed once the root is processed. SRCs are attempted only when no RCC is
violated and the relaxation is still fractional, and at most $64$ SRCs are
added.\footnote{The fractionality test only saves work. Since
\eqref{eq:mp_cover} is an inequality, an integral solution may serve a target
twice and so violate an SRC, although deleting the repeated visits repairs it
at no extra cost (Section~\ref{sec:reformulation}); separating SRCs would cut
it off. RCCs hold for every integral solution and need no such test.}

\subsection{Pricing Problem}
\label{sec:bprc_cg}

Solving RMP-$\mathcal{F}$ produces both a primal solution, containing
the $\theta_r$ values, as well as a solution to the dual problem of
RMP-$\mathcal{F}$. This dual contains a dual variable for each constraint
in RMP-$\mathcal{F}$. Let $\pi_{j}\ge 0$ denote the dual variable of the covering constraint
\eqref{eq:mp_cover} of target $j$, $\lambda\le 0$ the dual of the fleet
constraint \eqref{eq:mp_fleet}, $\mu_{S}\ge 0$ the dual of the rounded
capacity cut \eqref{eq:rcc} for $S\in\mathcal{S}$, and $\sigma_{C}\le 0$ the
dual of the subset-row cut \eqref{eq:src} for $C\in\mathcal{C}$.
The pricing problem seeks routes that could improve the RMP-$\mathcal{F}$
solution, and we use these dual variables to determine which routes to add to
$\Omega'$. In particular, let
\begin{equation}
\mu(e)\;=\;\sum_{S\in\mathcal{S}:\ e\in\partial S}\mu_{S}
\label{eq:arc_credit}
\end{equation}
be the RCC credit of arc $e$. The \emph{reduced cost} of a route
$r\in\Omega$ is
\begin{equation}
\bar{c}_{r}=c_{r}-\sum_{j=1}^{n} a_{jr}\,\pi_{j}-\lambda
-\sum_{e\sqsubseteq r}\mu(e)
-\sum_{C\in\mathcal{C}}\Big\lfloor\tfrac{1}{2}\textstyle\sum_{j\in C}
a_{jr}\Big\rfloor\sigma_{C}.
\label{eq:redcost}
\end{equation}
That is, the cost $c_r$ is reduced by a ``prize'' $\pi_j$ for each target
visited and a credit $\mu(e)$ for each arc traversed, and increased by a
penalty $-\sigma_C\ge0$ for every second target of a triple $C$ served.
To improve the RMP-$\mathcal{F}$ solution, we must add a route with negative
reduced cost \cite{b42}; if none respects $\mathcal{F}$, the solution is
optimal for LP-$\mathcal{F}$ with the same cuts. To ensure we find such a
route, we search for the one with the most negative reduced cost, i.e., we
solve the \textit{pricing problem}
\begin{equation}
\min\bigl\{\,\bar{c}_{r}:\ r\in\Omega,\ e\not\sqsubseteq r\ \forall e\in\mathcal{F}\,\bigr\}.
\label{eq:pricing}
\end{equation}
The RCC is \emph{robust}: $\chi_{r}(S)$ is a sum of per-arc indicators, so its
dual folds into arc costs,
$\sum_{S\in\mathcal{S}}\chi_{r}(S)\mu_{S}=\sum_{e\sqsubseteq r}\mu(e)$.
The SRC is \emph{non-robust}: its route coefficient
$\lfloor\tfrac12\sum_{j\in C}a_{jr}\rfloor$ is not, so $\sigma_{C}$ requires
extra state in the labeling algorithm (Section~\ref{sec:bprc_labeling}).

Each pricing call returns a batch of routes with negative reduced cost
(up to a cap of 200); all are added to $\Omega'$ and the
RMP-$\mathcal{F}$ is re-solved. As in \cite{b14}, a route is added to $\Omega'$ if
$\bar{c}_{r}<-\varepsilon$ (in our case $\varepsilon=10^{-5}$).

\subsection{Labeling Algorithm for Pricing Problem}
\label{sec:bprc_labeling}
The pricing problem \eqref{eq:pricing} is solved by a labeling algorithm similarly to
\cite{b14,b17}. A label $\ell=(\tau,\,\mathcal{V},\,t^{\textnormal{lb}},\,t^{\textnormal{ub}},\,\delta,\,\rho)$
represents a partial route $\Gamma$ from the depot to a current target $\tau$:
$\mathcal{V}$ is the set of targets visited so far, $t^{\textnormal{lb}}$ and $t^{\textnormal{ub}}$ are lower and
upper bounds on the earliest time the vehicle can meet
$\tau$ (Section~\ref{sec:bprc_dominance}),
$\rho=(\rho_{\ell,C})_{C\in\mathcal{C}}$, where
$\rho_{\ell,C}=\lvert \mathcal{V}\cap C\rvert\bmod 2$, holds one parity bit per SRC
(one $64$-bit word, as $\lvert\mathcal{C}\rvert\le64$), and
\begin{equation}
\delta=-\lambda-\sum_{j\in \mathcal{V}}\pi_{j}-\sum_{e\sqsubseteq\Gamma}\mu(e)
-\sum_{C\in\mathcal{C}}
\Bigl\lfloor\tfrac{1}{2}\lvert \mathcal{V}\cap C\rvert\Bigr\rfloor\sigma_{C}
\label{eq:delta}
\end{equation}
is its accumulated dual contribution. The label additionally stores the visiting order $(j_1,\dots,j_i)$, the lower bounds
$t^{\mathrm{lb}}_k$ of its earlier meetings, and the last meeting point $m_i$ of the
schedule \eqref{eq:ub_gap}-\eqref{eq:ub}, which will be required in Section~\ref{sec:bprc_dominance}.
The algorithm maintains a priority queue of labels keyed by
$t^{\textnormal{lb}}+\delta$, a lower bound on the reduced cost of the label's partial route. It starts from the single label
$(0,\emptyset,0,0,-\lambda,\mathbf{0})$ at the origin and repeatedly pops the
label with the smallest key and extends it.

Extending a label $\ell=(\tau,\mathcal{V},t^{\textnormal{lb}},t^{\textnormal{ub}},\delta,\rho)$ to a
target $j\notin\mathcal{V}$, allowed while $|\mathcal{V}|<Q$ and the arc
$e_{\tau j}$ is not forbidden by branching, creates the label
\begin{equation}
\ell'=\bigl(j,\ \mathcal{V}\cup\{j\},\ t^{\textnormal{lb}\prime},\ t^{\textnormal{ub}\prime},\
\delta',\ \rho'\bigr),
\label{eq:extend}
\end{equation}
where
\begin{itemize}
\item $t^{\textnormal{lb}\prime}$ and $t^{\textnormal{ub}\prime}$ are the bounds
\eqref{eq:lb} and \eqref{eq:ub} for the partial route $\Gamma$ followed by $j$;
\item $\delta'=\delta-\mu(e_{\tau j})-\pi_{j}-\sum_{C\ni j:\ \rho_{C}=1}\sigma_{C}$
subtracts the new arc's credit \eqref{eq:arc_credit} and the new target's
prize, and charges $-\sigma_{C}\ge0$ for each cut $C\ni j$ whose parity bit
returns to $0$, i.e., on every second visit to $C$ \cite{b20}, so that
$\delta$ always equals the dual part of \eqref{eq:redcost} accumulated
along $\Gamma$;
\item $\rho'$ is $\rho$ with the bit of every $C$ that contains $j$ flipped.
\end{itemize}
The new label is inserted into the queue unless discarded by one of the two
pruning devices below. Extending $\ell$ to the destination $D$ instead forms
the route $r=(\Gamma,n{+}1)$, whose $\delta$ is updated by the last arc's
credit only; its cost $c_r$ is computed by the conic program of Section~\ref{sec:bprc_socp}, its reduced cost \eqref{eq:redcost} exactly
$\bar{c}_{r}=c_{r}+\delta$, and is added to $\Omega'$ if
$\bar{c}_{r}<-\varepsilon$.

A \textit{continuation} of a label is a sequence of at most
$Q-|\mathcal{V}|$ pairwise distinct targets not in $\mathcal{V}$, and the
route obtained by appending a continuation and then $D$ to $\Gamma$ is a
\textit{completion} of the label.

Two pruning devices keep the search tractable; neither discards a label that
has a completion of negative reduced cost unless a coexisting label has a
completion of no greater reduced cost.
First, a \textit{completion bound}: a completion of a label with $|\mathcal{V}|$
visits appends at most $Q-|\mathcal{V}|$ targets and hence at most $Q-|\mathcal{V}|+1$ arcs
, so
it can collect at most the $Q-|\mathcal{V}|$ largest remaining positive prizes and at
most $\mu_{\max}=\max_{e\in \mathcal{E}}\mu(e)$ of RCC credit per appended arc, while
further SRC penalties only increase the reduced cost. Hence $t^{\textnormal{lb}}+\delta$
minus these prizes and credits is a lower bound on the reduced cost of every
route extending the label; if it is not below $-\varepsilon$, the label is
discarded. Second, the dominance criterion of the next subsection discards
labels judged unable to lead to a better route than a coexisting label.

\subsection{Continuous Bounds and Dominance}
\label{sec:bprc_dominance}
We say that a partial route $\Gamma$ dominates another partial route $\Gamma'$
if every sequence of targets appendable to $\Gamma'$ is also appendable to
$\Gamma$, and no extension of $\Gamma'$ can result in a reduced cost smaller
than the equivalent extension of $\Gamma$. In that case, $\Gamma'$ can be pruned
without the loss of optimality. Testing this condition directly is impractical,
so what follows is an efficiently testable sufficient condition for it.

\begin{figure}[t]
\centering
\includegraphics[width=\columnwidth]{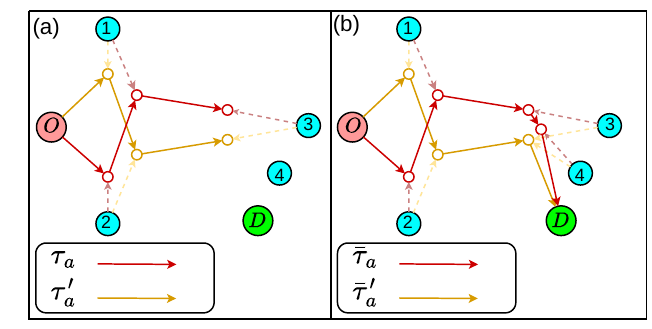}
\caption{(a) $\tau_a$ is an optimal (i.e. minimum vehicle travel time) trajectory visiting a partial route $\Gamma=(O,2,1,3)$, and $\tau_a'$ is an optimal trajectory visiting a partial route $\Gamma'=(O,1,2,3)$. The travel time of $\tau_a$ is equal to the travel time of $\tau_a'$. Since the set of visited targets is the same for both partial routes, a traditional dominance criterion could prune $\Gamma'$ assuming it is no better than $\Gamma$. (b) Appending the same sequence of targets to both partial routes results in an unequal cost increase. Travel time of $\bar{\tau}_a$ optimally traversing $\bar{\Gamma}=(O,2,1,3,4,D)$ is bigger than the travel time of $\bar{\tau}_a'$ optimally traversing $\bar{\Gamma}'=(O,1,2,3,4,D)$. This means that $\Gamma$ does not dominate $\Gamma'.$}
\label{fig:domcost}
\end{figure}

Dominance is complicated by the nature of floating targets:
In typical VRP scenarios, when appending targets to $\Gamma$, the cost incurred by visiting these
targets only depends on the final target in $\Gamma$. This means, that
appending the same sequence of targets to $\Gamma$ and
$\Gamma'$ that end on a common target would result in the same increase in cost. However, as
shown in Fig. \ref{fig:domcost}, when we have floating targets this is not the case. Appending the same sequence of
targets to $\Gamma$ and $\Gamma'$ actually increases the cost of $\Gamma$
more than for $\Gamma'$. This is because the cost of the sequence obtained by appending targets to $\Gamma$ depends
on the entire sequence of targets in $\Gamma$, not just its
final target.

Our criterion, inspired by the dominance rule of \cite{b14}, itself built on the bounds of \cite{b39},
resolves this with two closed-form bounds derived from the disk geometry.

\subsubsection{Bounds}
\begin{figure}[t]
\centering
\includegraphics[width=\columnwidth]{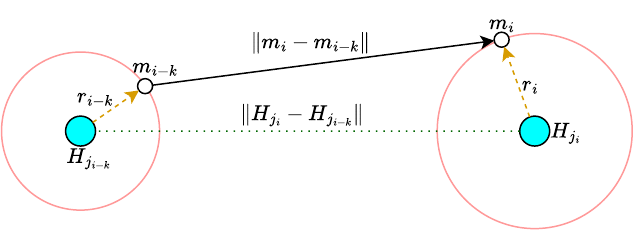}
\caption{Lower bound inequality between targets $j_{i-k}$ and $j_i$}
\label{fig:LB}
\end{figure}

For the partial route $\Gamma=(0,j_{1},\dots,j_{i})$ of a label, a feasible
schedule consists of meeting points $m_{i}$, with $m_{0}=O$, and meeting
times $\bar{t}_{i}=\sum_{i'\le i}t_{i'}$, cf.~\eqref{eq:rc_tgt}. We seek a
lower bound $t^{\textnormal{lb}}_{i}\le\bar{t}_{i}$ valid for every such schedule, with
$t^{\textnormal{lb}}_{0}=0$ at the origin. Since every target starts moving at $t=0$ and
must be at its meeting point when the vehicle arrives, the distance
$r_{i}=\lVert H_{j_{i}}-m_{i}\rVert$ from its home to its meeting point satisfies $r_{i}\le\bar{t}_{i}s^{\textnormal{tar}}_{j_{i}}$.
Each element $j_{i-k}$ of $\Gamma$ before $j_i$, $k=1,\dots,i$, yields a lower
bound on $\bar{t}_i$, obtained by ignoring the targets visited between
$j_{i-k}$ and $j_i$, and we take the maximum of these bounds. For $j_{i-k}$, the triangle inequality along
$H_{j_{i-k}}\!\to m_{i-k}\!\to m_{i}\!\to H_{j_{i}}$ (Fig.~\ref{fig:LB}) gives
\begin{equation}
\begin{split}
\lVert H_{j_{i-k}}-H_{j_{i}}\rVert
  &\le r_{i-k}+\lVert m_{i}-m_{i-k}\rVert+r_{i} \\
  &\le \bar{t}_{i-k}s^{\textnormal{tar}}_{j_{i-k}}
       +\bigl(\bar{t}_{i}-\bar{t}_{i-k}\bigr)s^{\textnormal{veh}}
       +\bar{t}_{i}\,s^{\textnormal{tar}}_{j_{i}}.
\end{split}
\label{eq:tri_chain}
\end{equation}
Rearranging \eqref{eq:tri_chain} to solve for $\bar{t}_{i}$ we get
\begin{equation}
\bar{t}_{i} \ \geq\
\frac{\lVert H_{j_{i-k}}-H_{j_{i}}\rVert+(s^{\textnormal{veh}}-s^{\textnormal{tar}}_{j_{i-k}})\,\bar{t}_{i-k}}
{s^{\textnormal{veh}}+s^{\textnormal{tar}}_{j_{i}}}.
\label{eq:lb_1}
\end{equation}
By Assumption~\ref{ass:speed} the coefficient $s^{\textnormal{veh}}-s^{\textnormal{tar}}_{j_{i-k}}$ is
nonnegative, so replacing the true, unknown $\bar{t}_{i-k}$ in \eqref{eq:lb_1}
by the smaller $t^{\textnormal{lb}}_{i-k}$ preserves the inequality; the right-hand side then
depends only on known quantities, and every $k$ thus supplies an independently
valid lower bound on the time of the $i$-th meeting. We take the best bound over
\emph{all} previously visited targets of the label:
\begin{equation}
t^{\textnormal{lb}}_{i}=\max_{1\le k\le i}
\frac{\lVert H_{j_{i-k}}-H_{j_{i}}\rVert+(s^{\textnormal{veh}}-s^{\textnormal{tar}}_{j_{i-k}})\,t^{\textnormal{lb}}_{i-k}}
{s^{\textnormal{veh}}+s^{\textnormal{tar}}_{j_{i}}},
\label{eq:lb}
\end{equation}
with $H_{j_{0}}=O$ and $s^{\textnormal{tar}}_{j_{0}}=0$. The stored values
$t^{\textnormal{lb}}_{0},\dots,t^{\textnormal{lb}}_{i-1}$ are carried in the label, so evaluating
\eqref{eq:lb} costs $O(i)$ time, bounded by $O(Q)$ since a partial route visits at most
$Q$ targets.

\begin{figure}[t]
\centering
\includegraphics[width=\columnwidth]{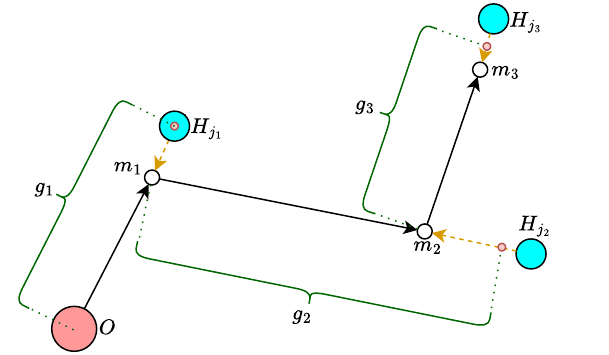}
\caption{Upper bound construction for a visiting sequence $(j_1, j_2, j_3)$.}
\label{fig:UB}
\end{figure}

The upper bound $t^{\textnormal{ub}}_{i}$ is the total time of an explicitly constructed
feasible schedule. Let $m_{0}=O$ and $t^{\textnormal{ub}}_{0}=0$, and suppose the vehicle
reaches the $(i{-}1)$-th meeting point $m_{i-1}$ of this schedule at time
$t^{\textnormal{ub}}_{i-1}$. Target $j_{i}$ walks from $t=0$ at full speed in a straight line
toward $m_{i-1}$, so by time $t^{\textnormal{ub}}_{i-1}$ the distance still separating it
from the vehicle is
\begin{equation}
g_{i}=\max\bigl(0,\ \lVert H_{j_{i}}-m_{i-1}\rVert-s^{\textnormal{tar}}_{j_{i}}\,t^{\textnormal{ub}}_{i-1}\bigr).
\label{eq:ub_gap}
\end{equation}
This distance closes at rate $s^{\textnormal{veh}}+s^{\textnormal{tar}}_{j_{i}}$ once the vehicle continues head-on
along the same line:
\begin{equation}
t^{\textnormal{ub}}_{i}=t^{\textnormal{ub}}_{i-1}+\frac{g_{i}}{s^{\textnormal{veh}}+s^{\textnormal{tar}}_{j_{i}}},
\label{eq:ub}
\end{equation}
with the new meeting point $m_{i}$ on the segment between $m_{i-1}$ and
$H_{j_{i}}$ (Fig.~\ref{fig:UB}). If the target arrives early ($g_{i}=0$), it simply waits at
$m_{i-1}$. Every step of this schedule is a feasible trajectory for both sides, so
$t^{\textnormal{ub}}_{i}$ is achievable and therefore a valid upper bound on the optimal time
of the $i$-th meeting. The upper-bound recursion extends a label in $O(1)$ time.

\subsubsection{Dominance criterion}
\label{sec:bprc_criterion}

Consider two labels $a=(\tau, \mathcal{V}_a, t^{\textnormal{lb}}_a, t^{\textnormal{ub}}_a, \delta_a, \rho_a)$ and $b = (\tau, \mathcal{V}_b, t^{\textnormal{lb}}_b, t^{\textnormal{ub}}_b, \delta_b, \rho_b)$ at the same target $\tau$ (home $H_\tau$, speed $v_\tau$) with $\mathcal{V}_a \subseteq \mathcal{V}_b$, representing partial routes $\Gamma$ and $\Gamma'$. We want to check whether $a$ dominates $b$. Let $\tilde{r}'$ be any completion of $b$ and let $\tilde{r}$ be the route obtained by appending the same continuation to $\Gamma$, with accumulated dual contributions $\tilde{\delta}'$ and $\tilde{\delta}$. Label $a$ dominates $b$ if $c_{\tilde{r}} + \tilde{\delta} \leq c_{\tilde{r}'} + \tilde{\delta}'$ for every such completion, i.e.,
\begin{equation}
\underbrace{c_{\tilde{r}'} - c_{\tilde{r}}}_{\text{cost advantage}} \;+\; \underbrace{\tilde{\delta}' - \tilde{\delta}}_{\text{dual advantage}} \;\geq\; 0.
\label{eqn:sum_of_advantages_geq_0}
\end{equation}
Neither advantage is known without fixing the completion, so we replace each by a lower bound that holds for every completion of $b$: the dual advantage is at least $\delta_b-\hat{\delta}_a$ and the cost advantage is at least $-\max\{M_{ab},\Pi_b M_{ab}\}$, with $\hat{\delta}_a$, $M_{ab}$ and $\Pi_b$ defined below. We therefore discard $b$ whenever
\begin{equation}
  \hat{\delta}_a \;+\; \max\bigl\{\,M_{ab},\; \Pi_b\,M_{ab}\,\bigr\} \;\le\; \delta_b .
  \label{eq:dominance_exact}
\end{equation}

\textit{Dual advantage.} Before any target is appended, the dual advantage is $\delta_b-\delta_a$. Appending the same targets to both labels adds the same prizes and arc credits, but an SRC $C$ whose parity bit is set in $\rho_a$ and not in $\rho_b$ may charge $a$ one penalty $-\sigma_C \ge 0$ that it does not charge $b$ and, by the parity argument of \cite{b20}, at most one. The dual advantage is therefore at least $\delta_b-\hat{\delta}_a$, where
\begin{equation}
  \hat{\delta}_a \;=\; \delta_a \;+ \sum_{\substack{C \in \mathcal{C} \\
  \rho_{a,C} = 1,\; \rho_{b,C} = 0}} (-\sigma_C).
  \label{eq:src-correction}
\end{equation}

\textit{Cost advantage.} Let $(m_b, t_b)$ be the meeting point and time of
$\tau$ in an optimal schedule of $\tilde{r}'$; $t_b \ge t^{\textnormal{lb}}_{b}$. We bound
how much later than $b$ the vehicle of $a$ can reach $m_b$; from there it can
serve the same remaining targets at the same meeting points as $b$. Label $a$
follows the explicit schedule of \eqref{eq:ub_gap}--\eqref{eq:ub}, meeting
$\tau$ at time $t^{\textnormal{ub}}_{a}$ at a point $m_a$, and then drives to $m_b$. Both
meeting points lie in the disk of radius $v_\tau t$ around $H_\tau$ at their
respective times, so
$\lVert m_a - m_b \rVert \le v_\tau (t^{\textnormal{ub}}_{a} + t_b)$
by the triangle inequality through $H_\tau$, and $a$'s vehicle reaches
$m_b$ no later than $A(t_b) = t^{\textnormal{ub}}_{a} + v_\tau (t^{\textnormal{ub}}_{a} + t_b)/s^{\textnormal{veh}}$.
Write
\begin{equation}
  M_{ab} \;=\; t^{\textnormal{ub}}_{a} \;+\;
  \frac{v_\tau\bigl(t^{\textnormal{ub}}_{a}+t^{\textnormal{lb}}_{b}\bigr)}{s^{\textnormal{veh}}} \;-\; t^{\textnormal{lb}}_{b}
  \;=\; A(t^{\textnormal{lb}}_{b}) - t^{\textnormal{lb}}_{b}
  \label{eq:margin}
\end{equation}
for the resulting time margin. By Assumption~\ref{ass:speed},
$v_\tau \le s^{\textnormal{veh}}$, so $(v_\tau/s^\textnormal{veh}-1) \leq 0$. This means that $A(t_b)-t_b = (v_\tau/s^\textnormal{veh}-1)t_b+C_A$, where $C_A$ is independent of $t_b$, is nonincreasing in $t_b$. This, combined with the fact that $t^\textnormal{lb}_b \leq t_b$, ensures that
$A(t_b)-t_b\le M_{ab}$ for every $t_b\ge t^{\textnormal{lb}}_{b}$.

Suppose first that $M_{ab} \ge 0$. If $a$'s vehicle arrives at $m_b$ after
$t_b$, it is late by at most
$L(t_b) = A(t_b) - t_b \le M_{ab}$ and can replay $b$'s continuation delayed by
$L(t_b)$: every target of the continuation
can reach its scheduled meeting point by the original time and wait there, so
delaying all remaining meetings is feasible and increases the completion time
by exactly $L(t_b)$. If instead it arrives no later than $t_b$, it waits at
$m_b$ and the delay is zero. Either way the delay is at most $M_{ab}$, so the
cost advantage is at least $-M_{ab}$.

Suppose now that $M_{ab} < 0$, so $a$'s vehicle may arrive at $m_b$ early.
Arriving early only guarantees that $a$ can wait at $m_b$ and
replay $b$'s continuation, finishing at the same time as $b$. It does not let
$a$ append that continuation any sooner, because the reachability disk of each
remaining target grows with time, so the targets may not be able to reach their
meeting points ahead of $b$'s schedule. What an early vehicle can still convert
into a saving is limited by how fast the targets it has yet to visit are
allowed to move. For a target $j$ let
\begin{equation}
  \gamma_j \;=\; \frac{s^{\textnormal{veh}}-s^{\textnormal{tar}}_{j}}{s^{\textnormal{veh}}+s^{\textnormal{tar}}_{j}} \;\in\; [0,1],
  \label{eq:gamma}
\end{equation}
which is $1$ for a stationary target and $0$ for a target whose speed limit
matches the vehicles'. Let
\begin{equation}
  q_b \;=\; \min\{\,Q-|\mathcal{V}_b|,\; n-|\mathcal{V}_b|\,\} ,
  \qquad \Pi_b \;=\; \prod_{j \in \mathcal{J}_b} \gamma_j,
  \label{eq:pi}
\end{equation}
where $q_b$ is the largest number of targets any continuation of $b$ can still visit and $\mathcal{J}_b$ collects the $q_b$ smallest values of $\gamma_j$ among the targets $j \notin \mathcal{V}_b$.
The proof of Proposition~\ref{prop:exact} shows that an early vehicle retains
at least the fraction $\Pi_b$ of its earliness $-M_{ab}$, so the cost advantage
is at least $-\Pi_b M_{ab}$. As $\Pi_b M_{ab}\le M_{ab}$ when $M_{ab}\ge 0$ and
$\Pi_b M_{ab}\ge M_{ab}$ otherwise, the bound $-\max\{M_{ab},\Pi_b M_{ab}\}$
used in \eqref{eq:dominance_exact} covers both cases.

\begin{proposition}
\label{prop:exact}
If \eqref{eq:dominance_exact} holds then $a$ dominates $b$; hence
Algorithm~\ref{alg:bpc} is exact.
\end{proposition}

\textit{Proof.} Fix a continuation of $b$: targets $j_{1},\dots,j_{p}$ with
$p \le q_b$ and $j_{k} \notin \mathcal{V}_b$, then $D$; since $\mathcal{V}_a \subseteq \mathcal{V}_b$, $a$ can
append it as well. Let $(m_{k},t_{k})$, $k=0,\dots,p$, be the meetings of a
cheapest such completion of $b$---with $(m_{0},t_{0})$, $t_{0}\ge t^{\textnormal{lb}}_{b}$,
its meeting with $\tau$---and $t_{p+1}$ its arrival at $D$, the route's cost.
As argued above, $a$'s vehicle can be at $m_{0}$ by
$A(t_{0})=t^{\textnormal{ub}}_{a}+v_\tau(t^{\textnormal{ub}}_{a}+t_{0})/s^{\textnormal{veh}}$; if $A(t_{0}) \ge t_{0}$ the
replay-and-delay argument applies unchanged, so assume $a$ is early by
$\Delta_{0}=t_{0}-A(t_{0})>0$.

We now prove by induction that $a$'s vehicle can serve $j_{1},\dots,j_{k}$ and stand at $m_{k}$ by
time $t_{k}-\Delta_{k}$, where $\Delta_{k}=\gamma_{j_{k}}\Delta_{k-1}$; the base case
$k=0$ is the assumption above. For the step, write $s=s^{\textnormal{tar}}_{j_{k}}$ and let the
vehicle, at $m_{k-1}$ at time $t_{k-1}-\Delta_{k-1}$, meet $j_{k}$ at the point
$p_{k}$ at distance $\beta=\Delta_{k-1}s\,s^{\textnormal{veh}}/(s^{\textnormal{veh}}+s)$ from $m_{k}$ on the
segment towards $H_{j_{k}}$ (if $\beta>\lVert m_{k}-H_{j_{k}}\rVert$, cap it
there; the arrival below is then only earlier). The target can be at $p_{k}$ by
$t_{k}-\beta/s$, since
$\lVert p_{k}-H_{j_{k}}\rVert=\lVert m_{k}-H_{j_{k}}\rVert-\beta\le
s\,t_{k}-\beta$; so can the vehicle: leaving $m_{k-1}$ at $t_{k-1}-\Delta_{k-1}$ it reaches
$p_{k}$ by $t_{k-1}-\Delta_{k-1}+\lVert m_{k-1}-p_{k}\rVert/s^{\textnormal{veh}}$, and
$\lVert m_{k-1}-p_{k}\rVert\le\lVert m_{k-1}-m_{k}\rVert+\beta\le
s^{\textnormal{veh}}(t_{k}-t_{k-1})+\beta$ bounds this by
$t_{k}-\Delta_{k-1}+\beta/s^{\textnormal{veh}}$, which equals $t_{k}-\beta/s$ since
$\beta(1/s+1/s^{\textnormal{veh}})=\Delta_{k-1}$ by the choice of $\beta$. Walking the remaining distance
$\beta$ to $m_{k}$, it stands there by
$t_{k}-\Delta_{k-1}+2\beta/s^{\textnormal{veh}}=t_{k}-\gamma_{j_{k}}\Delta_{k-1}$, proving the
claim.

Since $D$ is stationary, the vehicle finishes by $t_{p+1}-\Delta_{p}$, and
$\Delta_{p}=\Delta_{0}\prod_{k=1}^{p}\gamma_{j_{k}}\ge\Delta_{0}\Pi_b$ because
every $\gamma_j\le1$ and $\mathcal{J}_b$ collects the $q_b$ smallest factors available to
any continuation. This route collects the same prizes and arc credits as $b$'s
and, after the adjustment \eqref{eq:src-correction}, at least as much SRC credit, so discarding
$b$ is justified whenever $\hat{\delta}_a-\Delta_{0}\Pi_b\le\delta_b$, i.e.,
whenever $\hat{\delta}_a+\Pi_b\,M(t_{0})\le\delta_b$ with
$M(t)=t^{\textnormal{ub}}_{a}+v_\tau(t^{\textnormal{ub}}_{a}+t)/s^{\textnormal{veh}}-t$. Since $M$ is nonincreasing
($v_\tau\le s^{\textnormal{veh}}$ by Assumption~\ref{ass:speed}), $M(t^{\textnormal{lb}}_{b})=M_{ab}$
and $t_{0}\ge t^{\textnormal{lb}}_{b}$, we have $M(t_{0})\le M_{ab}$, hence
$\Pi_b\,M(t_{0})\le\Pi_b\,M_{ab}\le\max\{M_{ab},\Pi_b M_{ab}\}$, and
\eqref{eq:dominance_exact} implies the condition above. \hfill$\blacksquare$

Two special cases delimit \eqref{eq:dominance_exact}. If every remaining target
is stationary then $\Pi_b=1$ and the whole earliness is credited. If some
remaining target may move at $s^{\textnormal{veh}}$ then $\Pi_b=0$ and the rule
degenerates to $\hat{\delta}_a \le \delta_b$: no head start survives in that
regime. Between the two, $\Pi_b$ decays geometrically in
the number of remaining visits, so the credit is worth most near the end of a
route and little at its beginning. The test remains a single scalar comparison
per label pair: the $\gamma_j$ are sorted once per pricing call, and computing $\Pi_b$
costs one pass over that sorted list per label.

\subsubsection{A relaxed criterion}
\label{sec:bprc_relaxed}
Crediting the whole earliness regardless of the remaining targets, i.e.\
setting $\Pi_b=1$ in \eqref{eq:dominance_exact}, gives the relaxed criterion
\begin{equation}
  \hat{\delta}_a \;+\; t^{\textnormal{ub}}_{a} \;+\;
  \frac{v_\tau \bigl(t^{\textnormal{ub}}_{a} + t^{\textnormal{lb}}_{b}\bigr)}{s^{\textnormal{veh}}}
  \;\le\; \delta_b \;+\; t^{\textnormal{lb}}_{b} .
  \label{eq:dominance}
\end{equation}
Since $\max\{M_{ab},\Pi_b M_{ab}\}\ge M_{ab}$, \eqref{eq:dominance} discards
every label that \eqref{eq:dominance_exact} discards, and more; it also skips
the computation of $\Pi_b$. It is not a sufficient condition for dominance:
in the early case it lets $a$'s head start pay for a dual deficit
$\hat{\delta}_a > \delta_b$ that, as argued above, no continuation may be able
to recover. A pricing call using \eqref{eq:dominance} may therefore terminate
without having found an improving route that exists, so
Algorithm~\ref{alg:bpc} run with \eqref{eq:dominance} in place of
\eqref{eq:dominance_exact} is a heuristic. On every benchmark instance
with a certified optimum, however, it returned that optimum
(Section~\ref{sec:results}).

\subsection{Exact Route Costs}
\label{sec:bprc_socp}
The cost of a route ($c_r$) is the optimal value
of the convex program \eqref{eq:routecost}--\eqref{eq:rc_bnd}, which we solve with
the barrier method of an off-the-shelf conic solver. Pricing, however, may re-generate the same routes repeatedly as
the duals evolve across route-generation rounds. Thus, we store costs and meeting points for routes in a table, so they can be looked up when requested repeatedly for the same route.

\section{Computational Results}
\label{sec:results}

We evaluate the algorithm on the VRPFT benchmark of Gambella \textit{et al.}
\cite{b8} in both of its configurations: with the dominance criterion
\eqref{eq:dominance_exact} of Proposition~\ref{prop:exact} (exact) and with
the relaxed criterion \eqref{eq:dominance} (heuristic). The baselines are
the dynamic program of Zhang \textit{et al.} \cite{b13} and the published
results of Wang \textit{et al.} \cite{b41}; all 18 instances from \cite{b8}
are reported.

All runs were executed natively on Amazon EC2 \texttt{c8a} instances
(AMD EPYC ``Turin'', Zen~5, $4.5$\,GHz), under Ubuntu 24.04 with Gurobi 13.0.2 as
the conic and linear solver. Every run is single-threaded, pinned to a
dedicated physical core with \texttt{taskset}, and limited to $57{,}600$\,s
($16$ hours). The dynamic program was given an instance of the same processor and
clock. No implementation of \cite{b41} is available, so its values in
Table~\ref{tab:results} are as reported in that paper (1.61\,GHz Intel~i7,
CPLEX~12.7, $7{,}200$\,s limit).

\begin{table*}[t]
\caption{Objective value and running time on the VRPFT benchmark of Gambella
\textit{et al.} \cite{b8}. Instance ID is $n$-$K$-$Q$
(targets--vehicles--capacity). Values of \cite{b41} are as reported there
($7{,}200$\,s limit, different hardware and solver); $^{*}$: not proved
optimal. Boldface: fastest run among the identical-hardware experiments
(\cite{b13} and this work).}
\label{tab:results}
\begin{center}
\footnotesize
\setlength{\tabcolsep}{5pt}
\begin{tabular}{|l|r r|r r|r r r|}
\hline
& \multicolumn{2}{c|}{\textbf{Zhang \textit{et al.} \cite{b13}}}
& \multicolumn{2}{c|}{\textbf{Wang \textit{et al.} \cite{b41}}}
& \multicolumn{3}{c|}{\textbf{This work}} \\
\cline{2-8}
\textbf{Instance} & \textit{Obj.} & \textit{Time} & \textit{Obj.} & \textit{Time}
& \textit{Obj.} & \textit{Time} \eqref{eq:dominance_exact} & \textit{Time} \eqref{eq:dominance} \\
\hline
10-3-6 & 72.69 & 90.61       & 72.69        & 233.55        & 72.69 & 20.73        & \textbf{7.42} \\
10-4-5 & 62.16 & 26.07       & 62.14        & 19.56         & 62.14 & 11.00        & \textbf{7.83} \\
10-5-4 & 71.71 & 2.19        & 71.71        & 96.34         & 71.71 & 1.40         & \textbf{1.09} \\
12-3-6 & 76.92 & 705.87      & 76.91        & 115.01        & 76.91 & 47.92        & \textbf{14.90} \\
12-4-5 & 91.36 & 53.91       & 91.36        & 2{,}064.80    & 91.36 & 16.12        & \textbf{7.54} \\
12-5-5 & 80.08 & 37.49       & 80.08        & 738.97        & 80.08 & 17.81        & \textbf{5.99} \\
14-3-7 & 84.47 & 23{,}619.81 & 84.47        & 1{,}702.93    & 84.47 & 2{,}349.64   & \textbf{506.46} \\
14-4-6 & 97.62 & 1{,}484.67  & 97.62$^{*}$  & $>$7{,}200    & 97.62 & 172.80       & \textbf{27.82} \\
14-5-5 & 80.18 & 97.28       & 80.18        & 341.72        & 80.18 & 16.78        & \textbf{5.59} \\
16-3-8 & --    & $>$57{,}600 & 85.84        & 7{,}032.18    & 85.84$^{*}$  & $>$57{,}600  & \textbf{56{,}696.40} \\
16-4-6 & 76.71 & 4{,}983.58  & 83.20$^{*}$  & $>$7{,}200    & 76.71 & 1{,}539.49   & \textbf{460.41} \\
16-5-6 & 92.32 & 6{,}216.37  & 91.68        & 6{,}341.99    & 91.68 & 1{,}133.03   & \textbf{631.41} \\
18-3-8 & --    & $>$57{,}600 & 142.34$^{*}$ & $>$7{,}200    & --    & $>$57{,}600  & $>$57{,}600 \\
18-4-7 & --    & $>$57{,}600 & 87.57$^{*}$  & $>$7{,}200    & 80.54$^{*}$  & $>$57{,}600  & \textbf{42{,}025.80} \\
18-5-6 & 76.23 & 23{,}522.78 & 76.18        & 1{,}008.62    & 76.18 & 690.67       & \textbf{179.14} \\
20-3-9 & --    & $>$57{,}600 & 96.80$^{*}$  & $>$7{,}200    & --    & $>$57{,}600  & $>$57{,}600 \\
20-4-7 & --    & $>$57{,}600 & 139.59$^{*}$ & $>$7{,}200    & 110.08$^{*}$ & $>$57{,}600  & \textbf{6{,}404.57} \\
20-5-6 & --    & $>$57{,}600 & 126.99$^{*}$ & $>$7{,}200    & 126.99$^{*}$ & $>$57{,}600  & \textbf{16{,}571.50} \\
\hline
\end{tabular}
\end{center}
\end{table*}

\subsection{Objective values}
The exact configuration closed twelve instances (i.e. it terminated before the time limit on these instances); by
Proposition~\ref{prop:exact} these twelve values are certified optima. The
relaxed configuration closed the same twelve with identical objectives, so
the relaxed criterion \eqref{eq:dominance} lost no optimal solution on this
benchmark. It additionally finished on four of the remaining six instances
(16-3-8, 18-4-7, 20-4-7, 20-5-6); these values, marked $^{*}$ in
Table~\ref{tab:results}, carry no optimality certificate.

Measured against the certified optima, the dynamic program exceeds the
optimum on four of the twelve instances, by margins between $10^{-2}$ and
$0.64$, even though the objective values reported in \cite{b13} match those
of \cite{b8} exactly and the results in Table~\ref{tab:results} were obtained
with the implementation provided by the authors of \cite{b13}. Because
\eqref{eq:dominance_exact} certifies our values, these four figures are
suboptimal outputs of that implementation.

\subsection{Running times}
The exact configuration is faster than the dynamic program on all twelve
instances that both close, by a geometric mean factor of $5.3$ (smallest
margin $1.6\times$, on 10-5-4, largest $34\times$, on 18-5-6); even on the two
hardest instances, 14-3-7 and 16-4-6, certifying optimality takes
$2{,}349.64$\,s and $1{,}539.49$\,s against the dynamic program's
$23{,}620$\,s and $4{,}984$\,s. Each certified value is exact up to an
explicit tolerance of $K\varepsilon$ plus the conic solver's $10^{-8}$
optimality tolerance, about $5\cdot 10^{-5}$ in total. The relaxed
configuration is faster still, by a geometric mean factor of $14.7$ over the
dynamic program and by up to $131\times$ on 18-5-6 and $53\times$ on 14-4-6.

The speedup from relaxing the criterion---that is, the ratio of the last
two columns of Table~\ref{tab:results}---has a geometric mean of $2.8$ but is
far from uniform: it stays between $1.3$ and $3.0$ on the instances of
capacity at most $5$ and reaches $6.2$ on 14-4-6 and $4.6$ on 14-3-7. The
mechanism is visible in \eqref{eq:pi}: the credit $\Pi_b$ available to an
early label is a product of one factor $\gamma_j \le 1$ per remaining visit,
so it decays geometrically with the remaining capacity $Q-|\mathcal{V}_b|$.
On the benchmark instances the individual factors lie between $0.43$ and
$0.98$, which leaves a usable credit for labels near the end of a route and
almost none for the shallow labels where the label space is widest. The
relaxation therefore helps most on instances with larger capacity rather
than larger $n$; the deep instances 14-4-6 and 14-3-7 are the extreme cases.

Both configurations explore the same branch-and-bound tree, and that tree
explains where the time goes. Eleven of the twelve instances are closed at
the root node: the relaxation strengthened by the cuts \eqref{eq:rcc} and
\eqref{eq:src} matches the integer optimum, and no branching is required. Our
algorithm, in either configuration, branches only on 16-4-6, which it closes
after three nodes. Consequently the running time is essentially the root
node's, and inside it the split is extreme: the restricted master
re-optimizations total between $0.002$ and $0.29$\,s per instance, so pricing
accounts for more than $98\%$ of the running time on every instance.

\subsection{Comparison with the branch-and-price of \cite{b41}}

As \cite{b41} used different hardware, solver, and time limit, only objective
values are comparable: on all ten instances certified by both methods, the
values coincide. \cite{b41} additionally certifies $85.84$ on 16-3-8, which
our relaxed configuration reproduces independently within its limit. On
the five instances neither work closes, our relaxed
configuration attains the best known values on three: it improves 18-4-7 from
$87.57$ to $80.54$ and 20-4-7 from $139.59$ to $110.08$, and matches $126.99$
on 20-5-6.

\section{Conclusion}
\label{sec:conclusion}
We presented an exact branch-price-and-cut algorithm for the VRPFT whose pricing
subproblem is solved by a labeling algorithm with a new dominance criterion for
floating targets, together with a relaxed heuristic variant. On the benchmark of
\cite{b8}, the exact configuration certified twelve optima and was faster than the
dynamic program of \cite{b13} on each of them; the relaxed variant reproduced all
twelve at a fraction of the time, with a gain that grows with vehicle capacity.
The branch-and-price of Wang \textit{et al.}~\cite{b41} closes one further
instance, while we close two that it leaves open. Future work includes applying the dominance test separately to each possible next
target, so that individual extensions of a label can be pruned even when the label
itself is not dominated, and extending the criterion to targets with time windows
or release times.
\section*{Acknowledgment}
The authors thank Sivakumar Rathinam for helpful comments on this work.

\appendices
\section{Exhaustive Backtracking Search}
\label{app:backtracking}

Algorithm~\ref{alg:backtrack} is the backtracking search that
Algorithm~\ref{alg:bpc} calls to build the first routes and the incumbent
(lines~\ref{ln:init_start}--\ref{ln:init_end}) and to repair a node whose pool
cannot cover every target (lines~\ref{ln:repair_start}--\ref{ln:repair_end}).
Given a set $\mathcal{F}$ of forbidden arcs, with $\mathcal{F}=\emptyset$ at
the root and $\mathcal{F}=\mathcal{F}_N$ at node $N$, it looks for at most $K$
routes that together serve every target exactly once and use no arc of
$\mathcal{F}$. We call such a set of routes a \emph{partition respecting
$\mathcal{F}$}.

As in the depth-first construction of~\cite{b14}, routes are built
one at a time. A state of the search consists of the set $\mathcal{R}$ of
routes closed so far, the partial route $\Gamma=(0,j_1,\dots,j_q)$ under
construction, whose last element we denote by $\tau$, and the set $U$ of
targets not yet assigned to a route. The initial call is
\textsc{Backtrack}$(\emptyset,(0),\{1,\dots,n\})$. From a state, the search
first tries to append to $\Gamma$ each unassigned target $j$ whose arc
$e_{\tau j}$ is allowed
(lines~\ref{ln:bt_ext_start}--\ref{ln:bt_ext_end}), and only then tries to
close $\Gamma$ at the destination and open an empty route
(lines~\ref{ln:bt_close_start}--\ref{ln:bt_close_end}). If neither succeeds,
it returns to the previous state. The targets are tried in increasing order of
the distance $\lVert H_j-O\rVert$ from the origin to their homes; this order
is computed once and is the same in every state.

The search is purely combinatorial and stops at the first partition it finds.
Only then is the cost of each returned route computed by
\eqref{eq:routecost}--\eqref{eq:rc_bnd}; the routes are added to $\Omega'$
and their total cost replaces $z^*$ if it is smaller.

\begin{algorithm}[t]
\caption{\textsc{Backtrack}$(\mathcal{R},\Gamma,U)$}
\label{alg:backtrack}
\begin{algorithmic}[1]
\REQUIRE closed routes $\mathcal{R}$; partial route $\Gamma=(0,j_1,\dots,j_q)$
  ending at $\tau$; unassigned targets $U$; forbidden arcs $\mathcal{F}$
\ENSURE a partition respecting $\mathcal{F}$ that extends the state, or \textsc{fail}
\IF{$U=\emptyset$}
  \STATE \textbf{return} $\mathcal{R}\cup\{(\Gamma,n{+}1)\}$ if
    $e_{\tau,n+1}\notin\mathcal{F}$, else \textsc{fail}
\ENDIF
\IF{$q<Q$} \label{ln:bt_ext_start}
  \FORALL{$j\in U$ with $e_{\tau j}\notin\mathcal{F}$, nearest to $O$ first}
    \STATE $X\gets\textsc{Backtrack}(\mathcal{R},(\Gamma,j),U\setminus\{j\})$
    \IF{$X\neq\textsc{fail}$}
      \RETURN $X$
    \ENDIF
  \ENDFOR
\ENDIF \label{ln:bt_ext_end}
\IF{$q\geq 1$ \AND $e_{\tau,n+1}\notin\mathcal{F}$ \AND $|\mathcal{R}|+1<K$}
  \label{ln:bt_close_start}
  \RETURN $\textsc{Backtrack}(\mathcal{R}\cup\{(\Gamma,n{+}1)\},(0),U)$
  \label{ln:bt_close_end}
\ENDIF
\RETURN \textsc{fail}
\end{algorithmic}
\end{algorithm}

\textit{Lemma 1:} Algorithm~\ref{alg:backtrack} terminates. It returns a
partition respecting $\mathcal{F}$ if one exists, and \textsc{fail} otherwise.

\textit{Proof.} Every recursive call either moves a target from $U$ to
$\Gamma$ or closes a nonempty route, which can happen at most $n$ and $K-1$
times along a branch, and every state has at most $n+1$ successors, so the
search tree is finite. A returned set of routes is a partition respecting
$\mathcal{F}$: every arc of every route, including its first and last, was
tested against $\mathcal{F}$ before being added; a route receives a target
only while $q<Q$ and is closed only if $q\geq 1$; each target leaves $U$
exactly once; and line~\ref{ln:bt_close_start} keeps
$|\mathcal{R}|\leq K-1$ until the last route is closed. Conversely, let
$r_1,\dots,r_m$, $m\leq K$, be a partition respecting $\mathcal{F}$. Building
$r_1$ target by target, closing it, then building $r_2$, and so on, is a
sequence of moves available to the search. The search abandons a state only after trying all of its successors,
so it either reaches the end of this sequence or returns another partition
first. \hfill$\blacksquare$

Pruning a node at which the search fails preserves exactness, even though
\eqref{eq:mp_obj}--\eqref{eq:mp_bin} also admits solutions that serve a target
more than once. Some optimal solution $x$ serves every target exactly once
(Section~\ref{sec:reformulation}), and such an $x$ respects the forbidden arcs
of exactly one child at each branching (Section~\ref{sec:bprc_overview}), so
by Lemma~1 the search cannot fail at a node that contains it.

\bibliographystyle{IEEEtran}
\bibliography{references}

\end{document}